\documentclass[letterpaper, 10 pt, conference]{ieeeconf}  

\IEEEoverridecommandlockouts                              

\usepackage{amsmath}
\usepackage{amssymb}
\usepackage{ifpdf}
\usepackage{cite}
\usepackage[pdftex]{graphicx}
\usepackage[algoruled]{algorithm2e}
\usepackage{array}
\usepackage[caption=false,font=footnotesize]{subfig}
\usepackage{url}
\usepackage{multicol}
\usepackage{booktabs}
\title{\LARGE \bf
Concurrent Flow-Based Localization and Mapping in Time-Invariant Flow Fields*
}

\author{Zhuoyuan~Song$^{1}$,~\IEEEmembership{Member,~IEEE,} and Kamran Mohseni$^{2}$,~\IEEEmembership{Senior Member,~IEEE}
\thanks{*This work was partially supported by the Office of Naval Research, the National Science Foundation, and the University of Hawai`i at M\={a}noa.}
\thanks{$^{1}$Zhuoyuan Song was with the Department of Mechanical and Aerospace Engineering, University of Florida. He is currently with the Department of Mechanical Engineering, University of Hawai`i at M\={a}noa, Honolulu, HI, 96822 USA
        {\tt\small zsong@hawaii.edu}}%
\thanks{$^{2}$Kamran Mohseni is the William P. Bushnell Endowed Chaired Professor in the Department of Mechanical and Aerospace Engineering and the Department of Electrical and Computer Engineering. He is the director of the Institute for Networked Autonomous Systems, University of Florida, Gainesville, FL, 32611 USA
        {\tt\small mohseni@ufl.edu}}%
}

\begin{document}

\maketitle
\thispagestyle{empty}
\pagestyle{empty}

\begin{abstract}

We present the concept of concurrent flow-based localization and mapping (FLAM) for autonomous field robots navigating within background flows. Different from the classical simultaneous localization and mapping (SLAM) problem, where the robot interacts with discrete features, FLAM utilizes the continuous flow fields as navigation references for mobile robots and provides flow field mapping capability with in-situ flow velocity observations.  This approach is of importance to underwater vehicles in mid-depth oceans or aerial vehicles in GPS-denied atmospheric circulations. This article introduces the formulation of FLAM as a full SLAM solution motivated by the feature-based GraphSLAM framework.  The performance of FLAM was demonstrated through simulation within artificial flow fields that represent typical geophysical circulation phenomena: a steady single-gyre flow field and a double-gyre flow field with unsteady turbulent perturbations.  The results indicate that FLAM provides significant improvements in the robots' localization accuracy and a consistent approximation of the background flow field.  It is also shown that FLAM leads to smooth robot trajectory estimates.

\end{abstract}

\section{INTRODUCTION}

Persistent localization capability is an essential prerequisite for most mobile robot applications.  When external navigation aids such as the global positioning system (GPS) is not accessible, or the accuracy is insufficient, it is critical for the robot to fully utilize ambient features as potential navigational references. Recently, environmental feature-based navigation solutions, especially simultaneous localization and mapping (SLAM)~\cite{CadenaC:16a}, have enabled several commercialized robotic technologies including robot vacuum cleaners and autonomous driving vehicles.  Nonetheless, many important field robotic applications do not enjoy a similar ``luxury". For instance, hurricane-sampling unmanned aerial vehicles (UAVs) and many autonomous underwater vehicles (AUVs) operate in harsh fluid environments that lack conventional features for navigation.  These robots face unique localization challenges and demand for non-conventional solutions~\cite{PaullL:14a}.

Geophysical circulations, including atmospheric and ocean currents, are ever-present phenomena surrounding aerial and marine robots.  The movements of these dynamic events follow certain tractable principles that have been active topics of research for centuries~\cite{SchneiderT:06a,BuckleyMW:16a}. Inspired by this observation, we previously proposed the concept of utilizing background flows as navigational features~\cite{Song:17f}. A flow-aided navigation scheme was introduced to alleviate dead reckoning error of an AUV in long-term mid-depth missions where conventional navigational support is scarce. It has been demonstrated that the background flow field can provide valuable information for the localization of field robots. However, this approach builds upon the assumption that the flow field is predictable.

Motivated by these results, we present a concept called FLAM, \textit{i.e.},~flow-based localization and mapping. Similar to the classical SLAM algorithm, where a mobile robot concurrently constructs a map of its surroundings and localizes itself with respect to the resulting map, FLAM aims at bringing together the capabilities of flow field reconstruction and flow-aided navigation.  This paper focuses on a simplified case with time-invariant background flow fields, exemplifying the general type of scenarios where the time-dependency of the background flow dynamics is known. Different from the classical SLAM problems, FLAM utilizes dynamic field features and the robot's observations of the features (relative flow velocity) are velocity dependent.  We demonstrate a full-SLAM realization where the robot's trajectory and the background flow map are estimated through post-processing the measurements from the inertial navigation system and the relative flow velocity observations. 

\section{PREREQUISITES ON ROBOT LOCALIZATION}\label{sec:localization}
When a mobile robot utilizes external features to mitigate its localization error, an aided navigation system is established. Depending on the characteristics of the navigation features, the resulting aided navigation systems vary. The vast majority of the existing aided navigation schemes can be classified as the first category, which utilizes the static properties of navigation features, \textit{e.g.}~positions of the landmarks~\cite{DellaertF:99a,Durrant-WhyteH:06a,ClausB:15a}. Nevertheless, there often are situations where the robot does not encounter a sufficient number of, or any (especially in marine applications) such features for the navigation system to reference; hence static-feature-based navigation schemes become inapplicable. The second category utilizes the features' time-dependent properties, such as the locations of moving beacons~\cite{KussatNH:05a,FallonMF:10a}, or teammate robots~\cite{MourikisAI:06a,Song:13b}. However, the dynamics governing the changes in the states of these features are often not utilized. 

One type of dynamic, field features, which is of particular interest to many aerial and underwater robot applications, is the background flow fields surrounding the robots.  Geophysical flows, such as atmospheric and oceanic circulations, often persist in the navigation domains of compact UAVs and AUVs.  Although often chaotic and turbulent, these large-scale fluid flows contain tractable dynamic components that can be captured and predicted through numerical simulations in combination with limited observation data. General circulation models (GCMs), for instance, provide statistical information and predictions for describing the evolution of hurricanes or dominant ocean currents~\cite{ChassignetEP:09a}. The resolution and accuracy of these prediction models has been ever improving thanks to the fast progress in modern computation.  As shown in \cite{Song:17f}, it has become promising to utilize background flow fields as valuable navigation references in large-scale vehicle navigation problems when conventional alternatives fail.  One limitation of this approach is its dependency on pre-generated background flow velocity maps.  The motivation behind FLAM is to allow a robot to concurrent map the unknown flow field and perform flow-aided navigation with it.

\section{FLOW-BASED LOCALIZATION AND MAPPING}\label{sec:FLAM}
A major component of classical SLAM problems is the construction of a map of the physical world using a robot's exteroceptive perception inputs.  Different from conventional SLAM problems where the maps often consist of disjointed, finite landmarks, map features in FLAM are continuous fluid medium.  Besides, as it will become clear later, the observations in FLAM consist of a finite number of samples of the local flow velocity while the ideal map of the flow field is infinite dimensional.  Special treatments for the observation model in FLAM are required, and an appropriate representation of the flow field is crucial.

Several studies exist on incorporating dynamic features into classical SLAM formulation.  Nonetheless, it is often assumed that static features are present and dominate the map.  The dependency on static features in SLAM is because that the convergence of the SLAM algorithm relies on the correlation between robot state and landmark estimates, as well as the invariant correlation among all landmarks. When landmarks possess unknown mobility, the convergence of SLAM cannot be guaranteed.  Nevertheless, when the dynamics of the features is known to the robot, the motion model in the classical SLAM formulation can be augmented accordingly to track the changes in the states of the features and preserve inter-feature correlation properly.  This observation motivates our vision for FLAM. 

We demonstrate the concept of FLAM through the scenario with a time-invariant background flow field. A robot's states at time $k$ is denoted by $\mathbf{x}_k$.  A prescribed mesh grid with $N$ nodes is adopted to form a finite representation of the flow field with the flow velocity at the location of each node represented by $\mathbf{v}_i \; \forall \; i = 1, \cdots, N$. For simplicity of representation, we define vectors $\mathbf{y}_k \triangleq [\mathbf{x}_k^\top, \mathbf{v}_1^\top, \cdots, \mathbf{v}_N^\top]^\top$ and $\mathbf{y}_{0:k} \triangleq [\mathbf{x}_{0:k}^\top, \mathbf{v}_1^\top, \cdots, \mathbf{v}_N^\top]^\top$.  The robot is assumed to be equipped with a sensor that measures the changes in states, $\mathbf{u}_k$, and an ambient flow velocity sensor that provides relative flow velocity observations, $\mathbf{z}_k$.

The following assumptions are made to simplify our presentation of FLAM. (i) The robot's motion can be approximated as a Markov process such that $p(\mathbf{x}_k \,|\, \mathbf{x}_{0:k-1}, \mathbf{u}_k) = p(\mathbf{x}_k \,|\, \mathbf{x}_{k-1}, \mathbf{u}_k)$. (ii) The ambient flow velocity observations are mutually independent given the state of the robot such that $p(\mathbf{z}_{1:k} \,|\, \mathbf{x}_k) = p(\mathbf{z}_{k} \,|\, \mathbf{x}_k)$. (iii) The effect of the robot's motion on the background flow field is negligible. (iv) The robot's motion and observation result in normally distributed quantities. Assumptions (i) and (ii) are generally acceptable for many probabilistic robotics problems~\cite{ThrunS:05a}.  Assumption (iii) is also valid for small-scale UAV and AUVs within strong geophysical circulations.  The last assumption is generally not valid, especially for problems with high nonlinearity.  Its adversarial effects have been observed and formally studied in the SLAM literature~\cite{HuangS:16a}. However, for applications with accurate robot heading estimate (linear dynamics) and high-frequency feedback on the changes in the states (central-limit theory), errors due to this assumption are often trivial. 

FLAM seeks the posterior probability distribution $p(\mathbf{y}_{0:k} \,|\, \mathbf{u}_{1:k}, \mathbf{z}_{1:k})$. It has been well-established that this distribution can be computed recursively based on the knowledge of the system model and sensor inputs~\cite{ThrunS:05a}.  According to Bayes rule and Assumptions (i) and (ii), it can be decomposed as
\begin{align}
p(\mathbf{y}_{0:k} \,|\, & \mathbf{u}_{1:k}, \mathbf{z}_{1:k}) \nonumber \\
& = \eta \; p(\mathbf{y}_{0:k}) \; \prod_{1:k} p(\mathbf{x}_{k} \,|\, \mathbf{x}_{k-1}, \mathbf{u}_k) \; p(\mathbf{z}_k \,|\, \mathbf{y}_k).
\end{align}
The solution to FLAM  can be found by maximizing the posterior.  It is convenient to transform this problem to the minimization of the negative-log posterior
\begin{align}\label{eq:cost_1}
\mathcal{J}_\text{FLAM} &= -\log p(\mathbf{y}_{0:k} \,|\, \mathbf{u}_{1:k}, \mathbf{z}_{1:k}) = - c_0 - \log p(\mathbf{x}_0)  \nonumber \\ 
& -\sum_{1:k}\left[ \log p(\mathbf{x}_k \,|\, \mathbf{x}_{k-1}, \mathbf{u}_k) + \log p(\mathbf{u}_k \,|\, \mathbf{y}_k) \right].
\end{align}

A motion model can be defined based on the system dynamics of the robot.  It provides a means to compute posterior distribution $p(\mathbf{x}_k \,|\, \mathbf{x}_{k-1}, \mathbf{u}_k)$ by incorporating robot inertial navigation system (INS) measurements successively.  Different from the traditional formulation of a robot motion model in SLAM problems, we rewrite the motion model in a form that is more consistent with the observation model to be introduced as $\hat{\mathbf{u}}_k = f(\mathbf{x}_{k-1}, \mathbf{x}_{k})$. The observation model computes the theoretical relative flow velocity sensor measurements given current robot state estimate as $
\hat{\mathbf{z}}_k = h(\mathbf{y}_k)$. By applying Assumption (iv), we obtain the Gaussian representations for
\begin{align*}
&p(\mathbf{x}_k \,|\, \mathbf{x}_{k-1}, \mathbf{u}_k) \propto p(\mathbf{u}_k \,|\, \mathbf{x}_{k-1}, \mathbf{x}_k) = c_1 \cdot \nonumber \\
& \;\;\, \exp \left\{-\frac{1}{2} \left[ \mathbf{u}_k - f(\mathbf{x}_{k-1}, \mathbf{x}_{k}) \right]^\top {R}^{-1}_k \left[ \mathbf{u}_k - f(\mathbf{x}_{k-1}, \mathbf{x}_{k}) \right] \vphantom{-\frac{1}{2}}\right\},
\end{align*}
and
\begin{align*}
& p(\mathbf{u}_k \,|\, \mathbf{y}_k) = c_2 \exp \left\{-\frac{1}{2} \left[ \mathbf{z}_k - h(\mathbf{y}_k) \right]^\top {Q}^{-1}_k \left[ \mathbf{z}_k - h(\mathbf{y}_k) \right] \right\},
\end{align*}
where ${R}_k$ and ${Q}_k$ are the covariance matrices associated with the errors of the INS and relative flow velocity measurements, respectively. This leads to a quadratic cost function 
\begin{align}
&\mathcal{J}_\text{FLAM} = c + \sum_{1:k} \left\{\left[ \mathbf{z}_k - h(\mathbf{y}_{k}) \right]^\top {Q}^{-1}_k  \left[ \mathbf{z}_k - h(\mathbf{y}_{k}) \right]\right.\nonumber\\
& \qquad \left. \left[ \mathbf{u}_k - f(\mathbf{x}_{k-1}, \mathbf{x}_{k}) \right]^\top {R}^{-1}_k  \left[ \mathbf{u}_k - f(\mathbf{x}_{k-1}, \mathbf{x}_{k}) \right] \right\}.
\end{align}
Note that constant $c$ now contains the initial-state cost. 

It is clear now that FLAM can be formulated as a nonlinear least-squared estimation problem. The solution to this problem is the best-fit robot trajectory and flow velocity vectors at all node locations that minimize the errors associated with all sensor measurements
\begin{equation}
\mathbf{y}^*_{0:k} = \arg\min_{\mathbf{y}_{0:k}} \mathcal{J}_\text{FLAM}(\mathbf{y}_{0:k}).
\end{equation}
This formulation lends similarities to GraphSLAM formulations \cite{ThrunS:05a, GrisettiG:10a} that pertain to several popular SLAM applications. Several approaches have been proposed to solve this optimization problem \cite{KonoligeK:10a,KummerleR:11a,KaessM:12a}. One solution is to linearize $\mathcal{J}_\text{FLAM}$ at each time step using the robot's state estimate following the Gauss-Newton or Levenberg-Marquardt algorithm.  

We can define error functions
\begin{align}
\mathbf{e}^u_k = \mathbf{u}_k - f({\mathbf{x}}_{k-1:k}) \quad \text{and} \quad
\mathbf{e}^z_k = \mathbf{z}_k - h({\mathbf{y}}_k). \label{eq:err1}
\end{align} 
Both error terms can be linearized around given state estimates; the Jacobians associated with the motion and observation errors, $F_k$ and $H_k$, can be computed as
\begin{align*}
F_k &= \frac{\partial \mathbf{e}^u_k(\mathbf{x}_{k-1:k})}{\partial \mathbf{x}_{k-1:k}} \biggr\rvert_{\hat{\mathbf{x}}_{k-1:k}}, \quad
H_k &= \frac{\partial \mathbf{e}^z_k(\mathbf{y}_{k})}{\partial \{\mathbf{x}_{k}, \mathbf{v}_i\}} \biggr\rvert_{\{\hat{\mathbf{x}}_{k},\, \hat{\mathbf{v}}_i \}}.
\end{align*}
The cost function has a linearized form of
\begin{equation}
\mathcal{J}_\text{FLAM} \approx c + \frac{1}{2}\delta \mathbf{y}_{0:k}^\top\,\Omega\,\delta \mathbf{y}_{0:k} + \delta \mathbf{y}_{0:k}^\top \, \boldsymbol{\xi},
\end{equation}
where the information matrix $\Omega$ and the potential vector $\boldsymbol{\xi}$ can be built up by successively incorporating all sensor measurements according to Algorithm~\ref{alg:linearization}.
\vspace{-3mm}
\begin{algorithm}
	\KwData{$\hat{\boldsymbol{y}}_{0:k}$, $\mathbf{u}_{1:k}$, $R_{1:k}$, $\mathbf{z}_{1:k}$, $Q_{1:k}$}
	\KwResult{$\Omega$, $\boldsymbol{\xi}$}
	$\Omega = 0$, $\boldsymbol{\xi} = 0$   \tcc*[r]{Initialization}
	\For{all $\mathbf{u}_k$}{
		$\Omega \mathrel{+}= F_k^\top R^{-1}_k F_k$ and
		$\boldsymbol{\xi} \mathrel{+}= F_k^\top R^{-1}_k \mathbf{e}^u_k$ \tcc*[r]{For terms related to $\mathbf{x}_{k-1}$ and $\mathbf{x}_{k}$} 
	}
	\For{all $\mathbf{z}_k$}{
		$\Omega \mathrel{+}= H_k^\top Q^{-1}_k H_k$  and
		$\boldsymbol{\xi} \mathrel{+}= H_k^\top Q^{-1}_k \mathbf{e}^z_k$ \tcc*[r]{For terms related to $\mathbf{x}_{k}$ and four $\mathbf{v}_{i}$}
	}
	
	\caption{Linearization of  $\mathcal{J}_\text{FLAM}$}
	\label{alg:linearization}
\end{algorithm}
\vspace{-3mm}

When provided an initial guess for $\mathbf{y}_{0:k}$, an optimal increment $\delta \mathbf{y}_{0:k}$ that minimizes $\mathcal{J}_\text{FLAM}$ can be obtained by solving the linear equation system resulted from setting the derivative of $\mathcal{J}_\text{FLAM}$ to zero:
\begin{equation}\label{eq:ls}
\Omega \, \delta \mathbf{y}^* + \boldsymbol{\xi} = 0.
\end{equation}
Note that such a linear system is often overdetermined since the number of sensor measurements is typically more substantial than the total number of variables.  It is appropriate to solve this system as a solution to the corresponding linear least squares fitting problem. Moreover, $\Omega$ is a positive-definite, sparse matrix as in feature-based GraphSLAM. It is often convenient to solve \eqref{eq:ls} using indirect, iterative approaches such as conjugate gradient, especially when the number of sensor measurements is large. 

\section{2D IMPLEMENTATION}\label{sec:case}
We report two two-dimensional case studies of a robot performing FLAM within gyre-type flow fields.  In the first test case, a steady, single-gyre flow was considered as the background flow field.  This case is similar to localization in a vector field~\cite{Song:14b} but with the background flow field unknown, such as WiFi-based indoor localization~\cite{YassinA:17a}. The second test case uses a turbulent flow field consisting of a steady, double-gyre flow component and an unsteady, turbulent component generated by kinematic simulation.

\subsection{Flow Field Construction}
The double-gyre phenomena in large-scale ocean circulation are typical in the northern mid-latitude ocean basins.  The double-gyre flow model has an elegant closed-form stream function and has been widely used as a standard test case for fluid transportation studies~\cite{Shadden:05a}. 
We construct the turbulent flow component using kinematic simulation (KS)~\cite{FungJCH:92a}. KS models are non-Markovian-Lagrangian models for turbulent-like flow structures widely adopted in the investigation of particle dispersion or collision where kinetic interactions do not play a vital role in the analysis.  Despite the simple mathematical forms of KS, it provides an ideal tool for introducing small-scale eddies over a self-similar energy spectrum to a mean flow in this study.  Turbulent flow field generated by such a means has shown to be sufficient for our previous study on flow-aided navigation, where robot navigation performance evaluated in artificial flow fields showed good agreement with results from actual field test data~\cite{Song:17f}. The process of generating a turbulent flow field that has a Kolmogorov-like energy spectrum can be found in several pieces of literature~\cite{FungJCH:92a, Song:17f}.

\subsection{Motion Model}
We define the robot's state vector as $\mathbf{x}_k \triangleq [\boldsymbol{x}^\top_k, \boldsymbol{v}^\top_k, \psi_k]^\top \in \mathbb{R}^5$ where $\boldsymbol{x}$ denotes the robot's location, $\boldsymbol{v}$ denotes its linear velocity, and $\psi$ denotes the robot's heading.  For each grid node, we estimate the flow velocity at that location represented by $\mathbf{v}_i \in \mathbb{R}^2 \; \forall \; i = 1, \cdots, N$ in the inertial frame $\{n\}$. The robot is assumed to be equipped with an INS that measures body acceleration and angular velocity, $\mathbf{u}_k \triangleq [\mathbf{a}^\top_k, r_k]^\top \in \mathbb{R}^3$, in the body-fixed frame $\{b\}$, and ambient flow sensors that provide relative flow velocity observations, $\mathbf{z}_k \triangleq [\Delta v_x, \Delta v_y]^\top \in \mathbb{R}^2$, in $\{b\}$. 

For many aerial and underwater applications, direct measurements of body velocity in the inertial frame are difficult to obtain.  We consider the robot's linear velocity state $\boldsymbol{v}_k$ as a pseudo input in \eqref{eq:err1} such that $\mathbf{u}^+_k = [\hat{\boldsymbol{v}}^\top_k, \mathbf{u}^\top_k]^\top$.  Since the movement of underwater vehicles of interest is often slow, the motion of the robot can be modeled based on the first-order Euler method as
\begin{align}
\boldsymbol{v}_k &= (\boldsymbol{x}_k - \boldsymbol{x}_{k-1})/\Delta t + \boldsymbol{w}^v_k,\\
\boldsymbol{a}_k &= \mathbf{R}(\psi_k)(\boldsymbol{v}_k - \boldsymbol{v}_{k-1})/\Delta t + \boldsymbol{w}^a_k,\\
r_k &= (\psi_k - \psi_{k-1})/\Delta t + w^r_k,
\end{align}
where ``$\boldsymbol{w}_k^{*}$" are zero Gaussian motion noises with zero mean and covariance $R_k = \text{diag}(\boldsymbol{w}^v_k, \boldsymbol{w}^a_k, w^r_k)$. Here $\mathbf{R}(\cdot) \triangleq R^b_n(\cdot)$ is the rotational matrix from $\{n\}$ to $\{b\}$. 

\subsection{Observation Model}
\begin{figure*} 
	\vspace{-1mm}
	\centering
	\subfloat[\label{subfig:bilinear}]{%
		\includegraphics[width=0.2\linewidth]{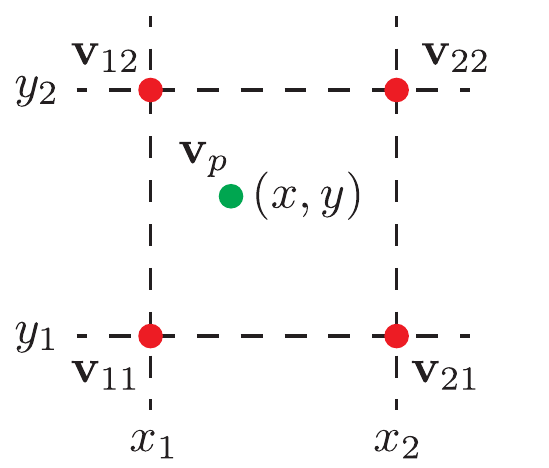}}
	\subfloat[\label{subfig:flowfield}]{%
		\includegraphics[trim={6mm 6mm 7mm 1mm}, clip, width=0.45\linewidth]{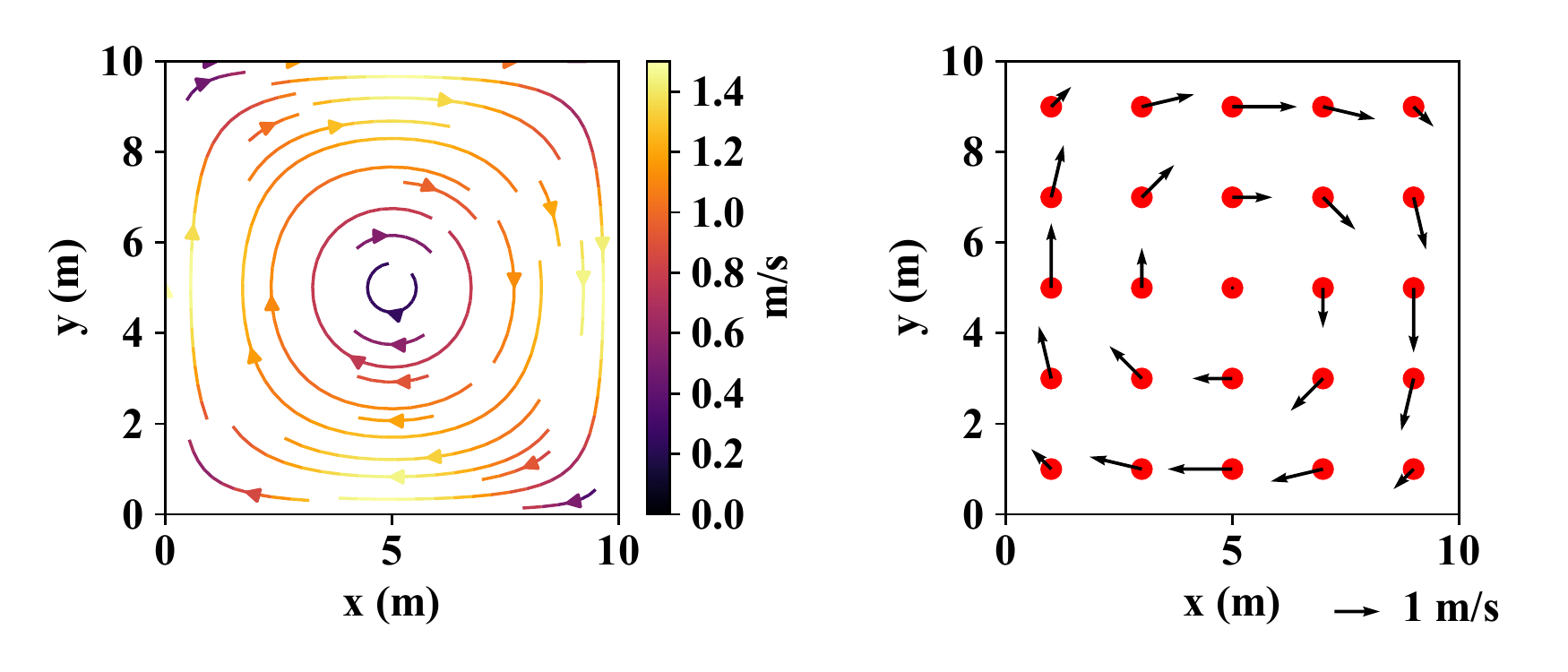}}
	\subfloat[\label{subfig:path}]{%
		\includegraphics[trim={7mm 6mm 7mm 1mm}, clip, width=0.32\linewidth]{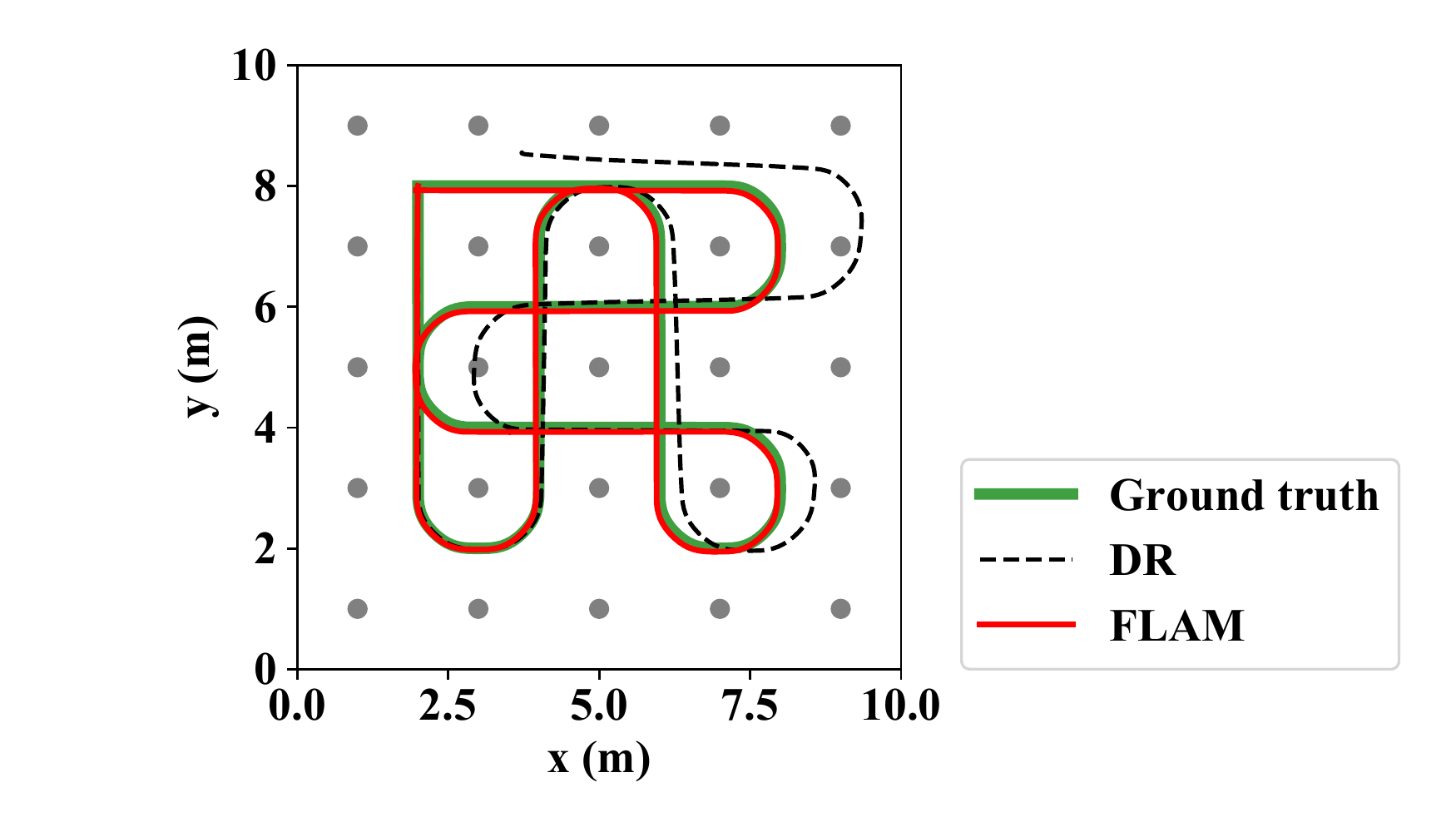}}
	\vspace{-0.1in}
	\caption{(a) The inertial background flow velocity at a given location $p(x, y)$ can be approximated through bilinear interpolation based on the flow velocities at the location of four cell nodes. (b) Streamline of the single-gyre flow field (left) and a mesh grid for a finite representation of the flow field (right). (c) Comparison of robot trajectories based on ground truth, dead-reckoning (DR), and estimation by FLAM. The vehicle started at (2~m, 8~m) heading south.}
	\vspace{-0.15in}
\end{figure*}

Given the inertial flow velocities at the vertices of a grid cell boxing the robot location, the inertial flow velocity at the robot's location can be computed through  interpolation.  The proper interpolation scheme can be chosen based on the complexity of the background flow field of interest and the desired accuracy.  For our case, bilinear interpolation was adopted.  As shown in Fig.~\ref{subfig:bilinear}, the inertial flow velocity, $\mathbf{v}_p$, at $\boldsymbol{x} = [x, y]$ can be calculated as
\begin{align}
\mathbf{v}_p &= \frac{y_2 - y}{y_2 - y_1}\frac{x_2 - x}{x_2 - x_1}\mathbf{v}_{11} + \frac{y_2 - y}{y_2 - y_1}\frac{x - x_1}{x_2 - x_1}\mathbf{v}_{21} \nonumber \\
& \, +\frac{y - y_1}{y_2 - y_1}\frac{x_2 - x}{x_2 - x_1}\mathbf{v}_{12} + \frac{y - y_1}{y_2 - y_1}\frac{x - x_1}{x_2 - x_1}\mathbf{v}_{22}.
\end{align}

\subsection{Simulation Setup}
We consider the case where the an AUV follows prescribed trajectories, where the maximum velocity of the robot is 2 m/s. Sample INS measurements were generated according to a consumer/industrial-grade system by corrupting the actual values with noises due to random walk and bias instability.  The relative flow velocity measurement sensor (observation sensor) was assumed to behave similarly to an acoustic Doppler current profiler (ADCP). The configurations of sensors are detailed in Table~I of \cite{Song:17f}.    

Assuming the initial robot states are known, an initial guess for $\mathbf{x}_{0:k}$ can be generated through dead-reckoning (DR) with INS measurements. By combining the result of DR with the ADCP measurements, an initial guess for $\{ \mathbf{v}_i \}$ can be obtained by solving the corresponding least squares fitting (LSF) problem.  Here we assume that the data association is known to the robot [Assumption~(iv)], \textit{i.e.},~the robot knows the indices of the four nodes boxing its actual location.  In addition, we solve the linear least squares problem \eqref{eq:ls} through conjugate gradient (CG) for this case study.  We adopted the damped version to avoid negative-definite $\Omega$ due to numerical error:
\begin{equation}\label{eq:damped}
(\Omega + \lambda I) \, \delta \mathbf{y}^* + \boldsymbol{\xi} = 0.
\end{equation}  
where we chose a constant damping factor $\lambda = 1e^{-3}$.  An adaptive damping scheme, such as the Levenberg-Marquardt algorithm, can also be applied to fine adjust the directional sensitivity of the iteration increment when necessary.  Spherical covariance was used for both $R$ and $Q$.  Convergence is often achieved after several iterations in our case.  Algorithm~\ref{alg:FLAM} summarizes the implementation of FLAM for this case study.
\vspace{-4mm}
\begin{algorithm}
	\KwData{$\hat{\boldsymbol{x}}_{0}$, $\mathbf{u}_{1:k}$, $R_{1:k}$, $\mathbf{z}_{1:k}$, $Q_{1:k}$, grid nodes}
	\KwResult{$\mathbf{y}^*_{0:k}$}
	
	$\mathbf{y}^{\text{init}}_{0:k} \leftarrow$ DR \& LSF   \tcc*[r]{Initial guess}
	\While{not converged}{
		$\Omega$, $\boldsymbol{\xi} \leftarrow$ Algorithm~\ref{alg:linearization} \\
		$\Omega = \infty$ at $\mathbf{x}_0$ \tcc*[r]{Anchor init. state}
		$\delta \mathbf{y}^* \leftarrow$ Solve~\eqref{eq:damped} through CG \\
		$\mathbf{y}_{0:k} \mathrel{+}= \delta \mathbf{y}^*$
	}
	
	\caption{FLAM}
	\label{alg:FLAM}
\end{algorithm}
\vspace{-4mm}

\subsection{Case I: Steady Single-Gyre Flow Field}
We first look at the case with a steady, single-gyre flow field.  Such a background flow velocity field has a single coherent structure and unique flow velocity vectors at each spatial location.  A streamline visualization of this flow field is shown in Fig.~\ref{subfig:flowfield}~(left).  We selected the left half of the double-gyre flow field and rescaled it to 10 m $\times$ 10 m by simply evaluating $(u,v)=(u(x/L,y/L,0),v(x/L,y/L,0))$, where the length-scale was chosen as $L=10$~m. A $5\times5$ grid mesh was defined to provide a finite representation of the flow field as shown in Fig.~\ref{subfig:flowfield} (right).  The inertial flow velocities at each node location constitute a map for FLAM. 

A comparison between DR and FLAM is shown in Fig.~\ref{subfig:path}.  The estimation result from FLAM shown here was obtained after five iterations of Algorithm~\ref{alg:FLAM}. It can be noticed that FLAM provides consistent state estimates in steady background flow fields without velocity field ambiguity, meaning that flow velocity vectors to spatial locations are injective.  

\begin{figure}
	\vspace{0mm}
	\centering
	\includegraphics[trim={20 0 15 20mm}, clip, width=1\linewidth]{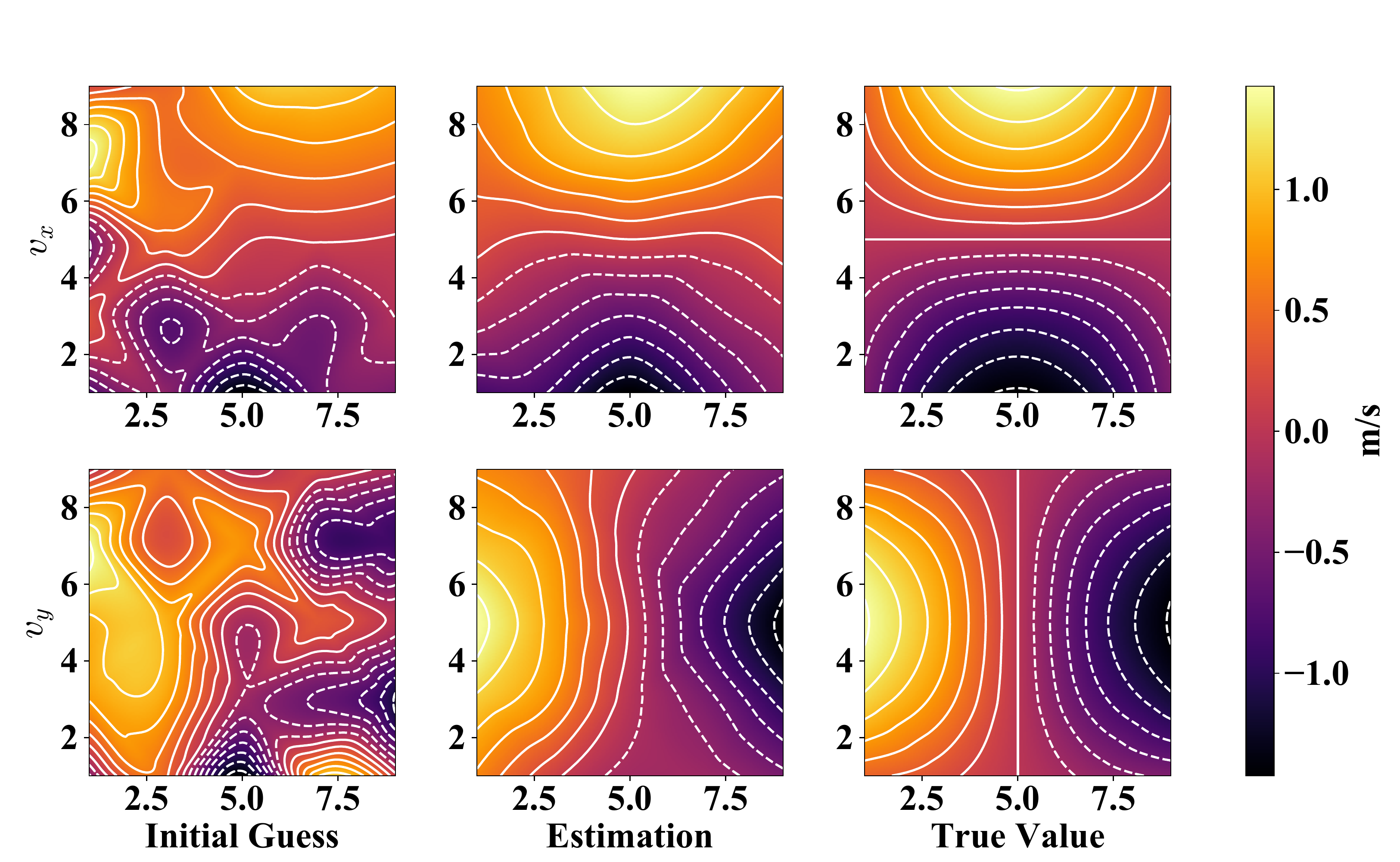}
	\vspace{-8mm}
	\caption{Evaluation of the flow mapping performance based on $x$ (top) and $y$ (bottom) velocity components in the single-gyre flow field. The figure shows the initial guess based on dead-reckoned robot trajectory and noisy ADCP observations (left), estimation results of FLAM (middle), and interpolation results using actual flow velocity vectors at node locations (right). }
	\label{fig:flow}
	\vspace{-6mm}
\end{figure}

Similar to classical SLAM, the convergence of FLAM can only be achieved with consistent estimation of both robot states and the background flow velocity map.  Fig.~\ref{fig:flow} summarizes the flow field mapping performance of FLAM.  The initial guess (left), as a result of the LSF of ADCP observations based on the dead-reckoned robot trajectory, and the FLAM estimation (middle) are compared to the interpolation results (right), based on actual flow velocity vectors at node locations, for both the $x$ (top) and $y$ (bottom) axial velocity components of the background flow.  All results shown here are interpolated based on node values through cubic splines. FLAM successfully captured the dominant flow field signatures although the initial guesses contained large error. The flow field reconstruction error is generally larger near the boundary region of the convex hull spanned by the mesh grid.  This is because those boundary nodes are rarely revisited by the robot, therefore, are less correlated with the rest of the flow map.  

\subsection{Case II: Turbulent Double-Gyre Flow Field}
Many flow fields that robots encounter in reality are often time-varying and turbulent.  We further evaluate the efficacy of FLAM in a turbulent flow field consisting of a steady, large-scale double-gyre component and a series of unsteady, small-scale turbulent components.  Although sharing similarities, this flow field differs from the steady single-gyre flow field in multiple ways. Firstly, the double-gyre flow component has two separated coherent structures, \textit{i.e.},~two counter-rotating gyres, creating a vertical transportation barrier at the center of the domain~\cite{Shadden:05a}.  In other words, the flow field on the left half is not correlated with the right half.  In addition, the turbulent flow component introduces time-dependency to the background flow field, the dynamics of which is not known to the robot.  The effect of this is similar to a hypothetical case of classical SLAM when the landmarks are allowed to have bounded location variations.  It can be hypothesized that this will increase the error residue of robot state estimate and may cause the FLAM to diverge as the magnitude of the turbulent effect approaches that of the large-scale component. 

\begin{figure}
	\centering
	\includegraphics[trim={20mm 0mm 12mm 79mm}, clip, width=0.85\linewidth]{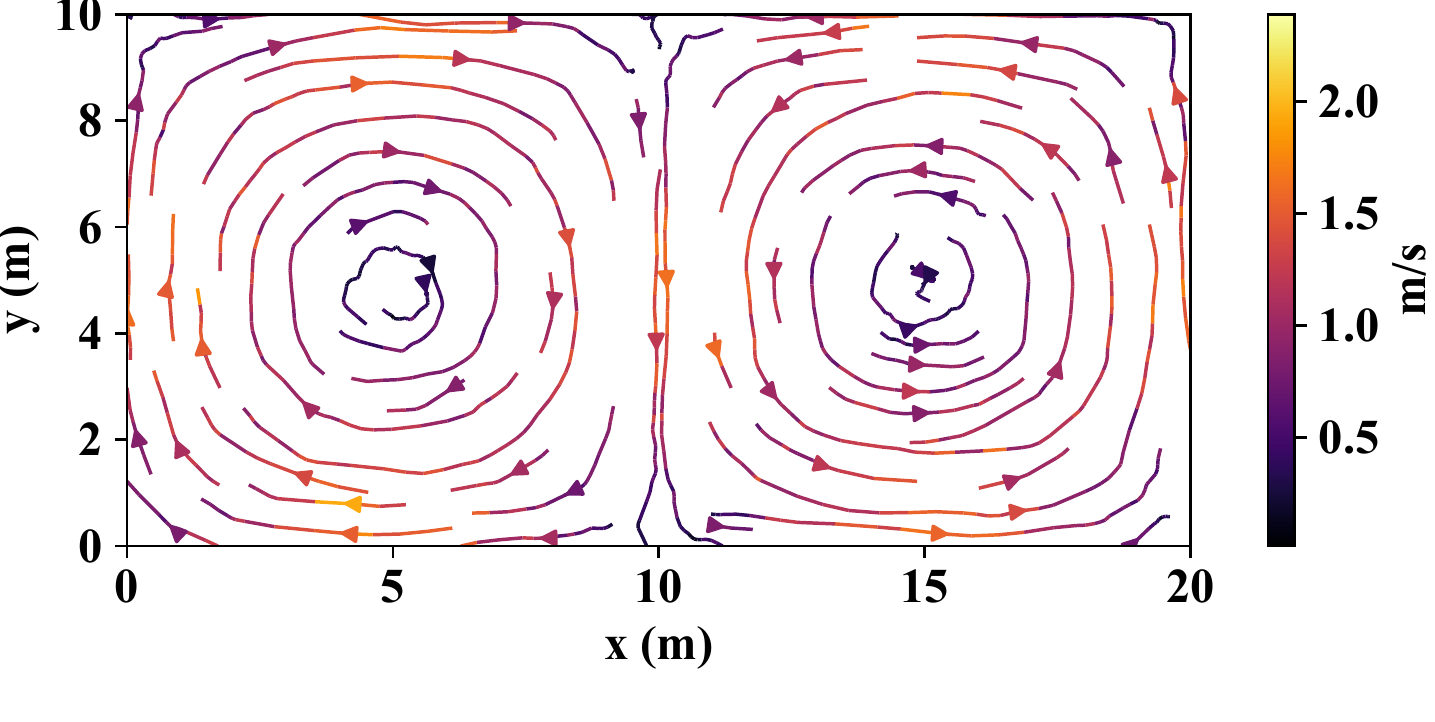}
	\vspace{-7mm}
	\caption{The turbulent flow field consisting of a steady, large-scale double-gyre component and a series of unsteady, small-scale turbulent components with a Kolmogorov-like energy spectrum.}
	\label{fig:flowfield_turbulent}
	\vspace{-2mm}
\end{figure}

\begin{figure}
	\centering
	\includegraphics[trim={15 0 10 18mm}, clip, width=1\linewidth]{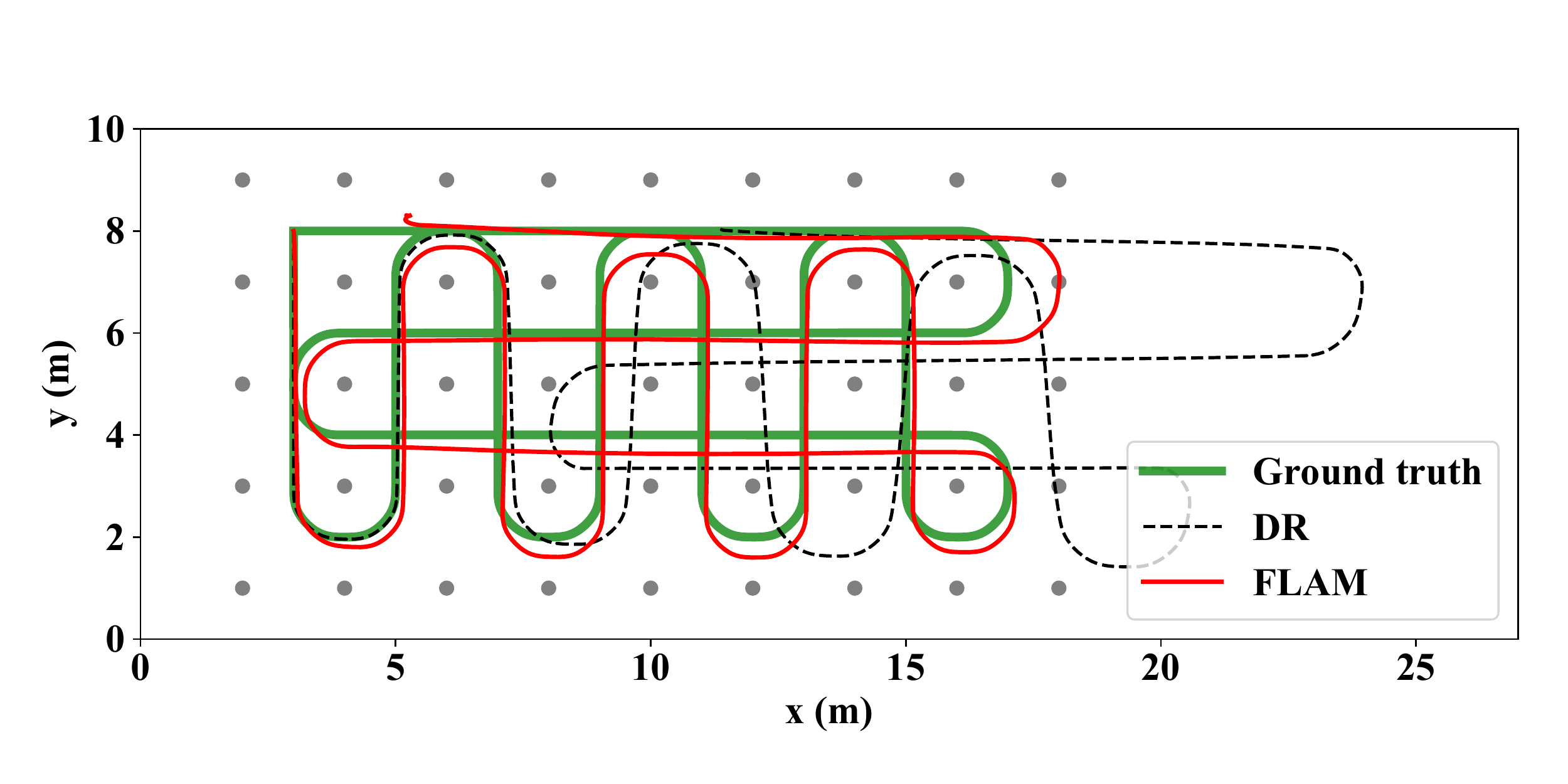}
	\vspace{-11.5mm}
	\caption{Comparison of robot trajectories based on ground truth, dead-reckoning (DR), and estimation by FLAM  in the turbulent, double-gyre flow field.. The mission started from (3~m, 8~m) heading south. Grid nodes indicate location of flow velocity the robot estimates as a finite representation of the entire flow field. }
	\label{fig:path_turbulent}
	\vspace{-4mm}
\end{figure}

\begin{figure}
	\centering
	\includegraphics[trim={0 0 0 8mm}, clip, width=0.8\linewidth]{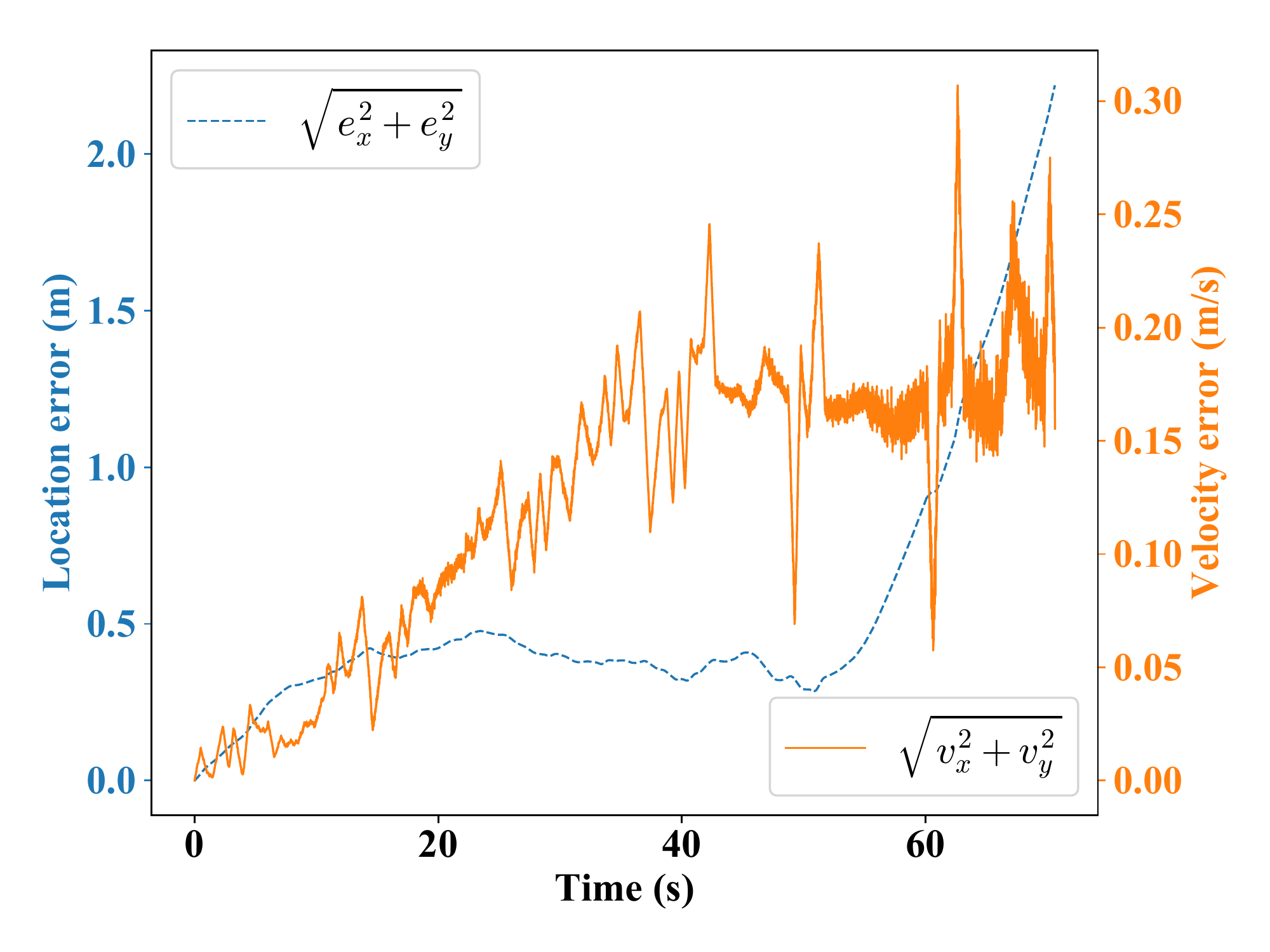}
	\vspace{-6mm}
	\caption{Estimation errors in robot location and linear velocity of FLAM in the turbulent, double-gyre flow field.}
	\label{fig:error_turbulent}
	\vspace{-2mm}
\end{figure}

The steady flow component was created based on the double-gyre model at $t=0$ with a lengthscale of 10 m.  The turbulent components were generated to have the large (internal) scale $L = 1$~m and the small (Kolmogorov) scale $\eta = 1e^{-3}$~m, resulting in a Reynolds number of $10^3$ ($Re = L/\eta$).  The turbulent double-gyre flow field at $t=0$ is shown in Fig.~\ref{fig:flowfield_turbulent}.  A $5\times9$ grid mesh was defined as a finite representation of the flow field shown in Fig.~\ref{fig:path_turbulent}, where the trajectory estimates are presented.  As expected, FLAM provides a significant improvement over the localization accuracy compared to DR.  A more substantial residual error, as speculated previously, can also be noticed due to the presence of unsteady turbulent flow components.  It is also observed that the localization error has a more dramatic increase towards the end of the mission.  To gain more insight into its cause, we analyze the root mean squared error of the localization and velocity estimates as the navigation proceeds.  As shown in Fig.~\ref{fig:error_turbulent}, the robot's velocity estimate first increases as the navigation proceeds and reaches a steady mean value that is approximately the average turbulent flow velocity.  As also observed in the previous test case, the velocity estimation error affects the localization error in a delayed fashion and causes large location deviation towards the latter portion of the mission. 

\begin{figure}
	\centering
	\includegraphics[trim={10mm 10mm 22mm 20mm}, clip, width=0.75\linewidth]{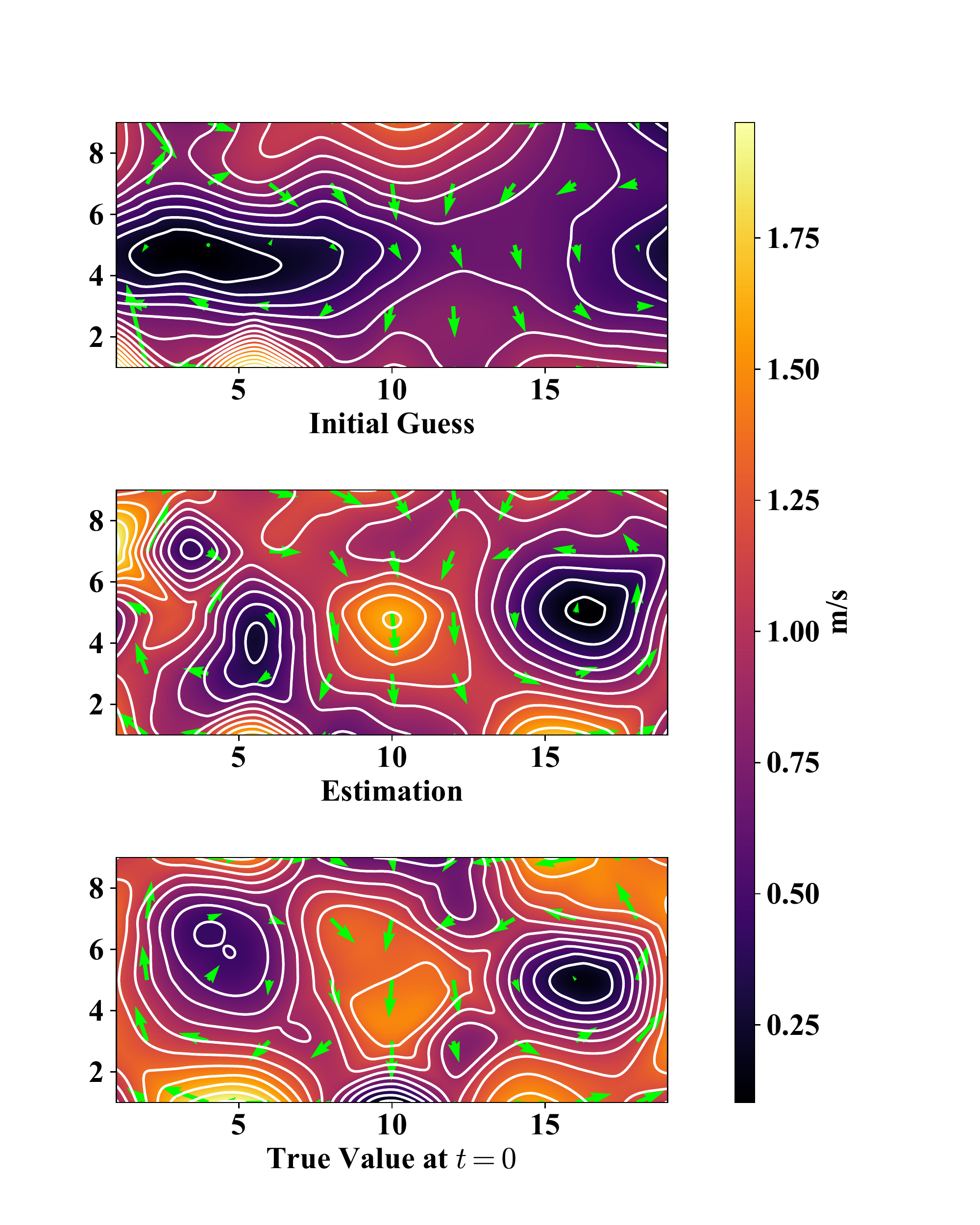}
	\vspace{-4mm}
	\caption{Flow field mapping performance evaluation in the turbulent, double-gyre flow field.  Figure shows the initial guess based on dead-reckoned robot trajectory and noisy ADCP observations (top), estimation results of FLAM (middle), and interpolation results using actual flow velocity vectors at node locations for $t=0$ (bottom).}
	\label{fig:flow_turbulent3}
	\vspace{-2mm}
\end{figure}

The presence of unsteady turbulent flows also has an evident impact on the mapping performance.  As shown in Fig.~\ref{fig:flow_turbulent3}, where the initial guess, FLAM estimate, and interpolation result with true node values are compared, LSF has a large error in capturing the dominant features in the flow field.  On the other hand, FLAM captures the underlying double-gyre flow structure although with comparably larger estimation inaccuracy compared to the previous case due to the unsteady turbulent effect.  Such a capability is valuable in studying coherent structures in geophysical circulations~\cite{HallerG:15a} with mobile robots.  

\section{CONCLUDING REMARKS}\label{sec:conclusion}
We presented the concept of concurrent flow-based localization and mapping (FLAM). Inspired by the classical SLAM formulation, FLAM provides a novel solution to localization problems faced by field mobile robots navigating within GPS-denied, landmark-deficit environments with nontrivial background flows.  We showed that the proposed FLAM scheme provides improved robot localization performance and consistent mapping of the background flow field in a steady, single-gyre flow field and a turbulent, time-invariant double-gyre flow field.  We hope this study will sheds some light on the value of background flow fields as navigation features in situations when conventional alternatives fail. 
\vspace{-1mm}

\addtolength{\textheight}{-12cm}   









\bibliographystyle{IEEEtran}
\bibliography{RefA2,bib/ran}

\end{document}